\documentclass{article} 
\usepackage{iclr2026_conference,times}
\usepackage{graphicx}
\usepackage{hyperref}
\usepackage{amsmath,amsfonts}
\usepackage{array}
\usepackage[caption=false,font=normalsize,labelfont=sf,textfont=sf]{subfig}
\usepackage{textcomp}
\usepackage{stfloats}
\usepackage{url}
\usepackage{verbatim}
\usepackage{tabularx}
\usepackage{amssymb}
\usepackage{multirow}
\usepackage{caption}
\usepackage{booktabs}
\usepackage{colortbl}
\usepackage{amsmath}
\usepackage{color}
\usepackage{bbding}
\usepackage{wrapfig}
\usepackage{balance} 
\usepackage[table]{xcolor}
\usepackage[most]{tcolorbox} 
\usepackage{setspace}

\definecolor{warmback}{RGB}{254, 252, 240}   
\definecolor{warmframe}{RGB}{225, 215, 190}  
\definecolor{warmtitle}{RGB}{140, 60, 40}    
\definecolor{oursburgundy}{HTML}{DCEEFF}
\newtcolorbox{querybox}{
    colback=warmback,
    colframe=warmframe,
    coltext=black,
    boxrule=0.6pt,
    arc=0pt, 
    left=10pt,              
    right=5pt,             
    top=8pt,                
    bottom=8pt,             
    enhanced,               
    breakable,               
    before upper={\setstretch{1.1}}
}

\title{DSPO: Diversity-aware Subjective Policy Optimization for Robust Emotional Reasoning}

\author{Cheng Ye$^1$, Weidong Chen$^1$, Bingyan Xu$^1$, Zhendong Mao$^1$ \\
$^1$University of Science and Technology of China, Hefei \\
\texttt{chenweidong@ustc.edu.cn}
}

\iclrfinalcopy 
\begin{document}

\maketitle

\begin{abstract}
Reinforcement Learning has significantly advanced the complex reasoning capabilities of Multimodal Large Language Models (MLLMs). However, prevailing RL algorithms, such as Group Relative Policy Optimization (GRPO), suffer a severe failure in emotion reasoning tasks. These methods heavily rely on deterministic hard-label supervision and point-wise isolated evaluation, creating a fundamental gap with the inherently subjective and continuously distributed nature of human emotions. Furthermore, unlike explicit physical objects, emotional states are deeply implicit within visual cues. This abstract nature exacerbates visual hallucinations in MLLMs, leading to plausible yet ungrounded emotional evidence.
To address these limitations, we propose \textbf{Diversity-Aware Subjective Policy Optimization (DSPO)}, a reinforcement learning framework that jointly promotes subjective affective coverage and visual grounding. First, we construct a context-grounded emotional distribution prior in the VAD space by combining the lexical prior of the annotated emotion with image-specific contextual information. Based on this prior, we introduce a Distribution-Aligned Emotional Diversity Reward (DEDR), which measures the leave-one-out marginal contribution of each candidate emotion within a rollout. DEDR rewards candidates whose inclusion brings the predicted affective set closer to the context-grounded prior, thereby preserving plausible subjective interpretations without encouraging unconstrained dispersion. We further develop Counterfactual Visual Intervention Gating (CVIG), which masks the visual region highlighted in the reasoning process and uses the resulting candidate-wise probability changes to reduce the weights of interpretations unsupported by visual evidence.
Extensive experiments demonstrate that DSPO achieves state-of-the-art performance across multiple public benchmarks, especially on the cross-domain performance, \emph{i.e.,} improving +10.8\% on average cross-domain accuracy than EMO-R3.\footnote{Code will be released in the final version of the paper.}
\end{abstract}

\section{Introduction}
Multimodal Large Language Models (MLLMs) have recently demonstrated remarkable proficiency in general-oriented perception and reasoning tasks ~\citep{yang2026attention,liu2025oryx,yue2025mllm,huang2025graph}. However, they frequently failed when applied to emotional computing and human-centered reasoning ~\citep{xie2024emovit,zhang2025mme,dingdongemotionthinker}. Unlike object-centric tasks~\citep{yao2026lens,zhang2026thinking,gu2026thinkmorph}, where ground truths are deterministic and visually explicit, emotion states are inherently latent, highly subjective, and deeply embedded within nuanced multimodal contexts. Evaluating emotion states requires models to go beyond superficial pattern recognition to capture micro-level expressions, complex social dynamics, and subtle psychological signals ~\citep{qin2026humansense,wang2026multi,yuan2026video}. Consequently, bridging the semantic gap between explicit visual elements and implicit human emotions remains a grand challenge for current MLLMs ~\citep{guo2025emoverse,chen2026face,yao2026adapt}.

Recently, Reinforcement Learning (RL), particularly Group Relative Policy Optimization (GRPO)~\citep{shao2024deepseekmath}, has emerged as a promising paradigm to elicit advanced reasoning capabilities in MLLMs~\citep{chaubeyavere,ge2026expand,chen2021cascade,chen2022multi}. Despite its success in deterministic tasks like mathematical reasoning, applying GRPO directly to emotion reasoning reveals two intrinsic limitations.
1) \textbf{Exacerbated Visual Hallucination.} When inferring implicit affective cues, the tendency of MLLMs to generate visual hallucinations is significantly exacerbated. Specifically, models frequently generate non-existent visual evidence to forcibly justify a plausible emotional conclusion. More critically, in the absence of objective physical anchoring, existing methods~\citep{fang2026emo} attempt to correct evidence-image inconsistencies through self-reflection mechanisms, which lacks external verification and easily fall prey to confirmation bias within MLLMs. This closed-loop self-justification not only amplifies reasoning errors but also reinforces hallucinated reasoning trajectories. Ultimately, this causes MLLMs to merely adopt shortcut learning to accommodate target labels without acquiring causal affective reasoning capabilities, thereby severely compromising their generalization and robustness in open-world scenarios.
2) \textbf{Sparse Discrete Reward.} Existing RL-based methods rely on discrete emotion labels as reward signals. This rigid matching mechanism fundamentally conflicts with the continuous and distributed nature of human emotion, where diverse subjective interpretations could naturally coexist. Penalizing valid minority perspectives inevitably forces the policy into mode collapse, causing it to merely fit the label distribution of the given dataset rather than learning the true emotional semantic space.

\begin{figure*}[t]        
\center{\includegraphics[width=0.95\linewidth] {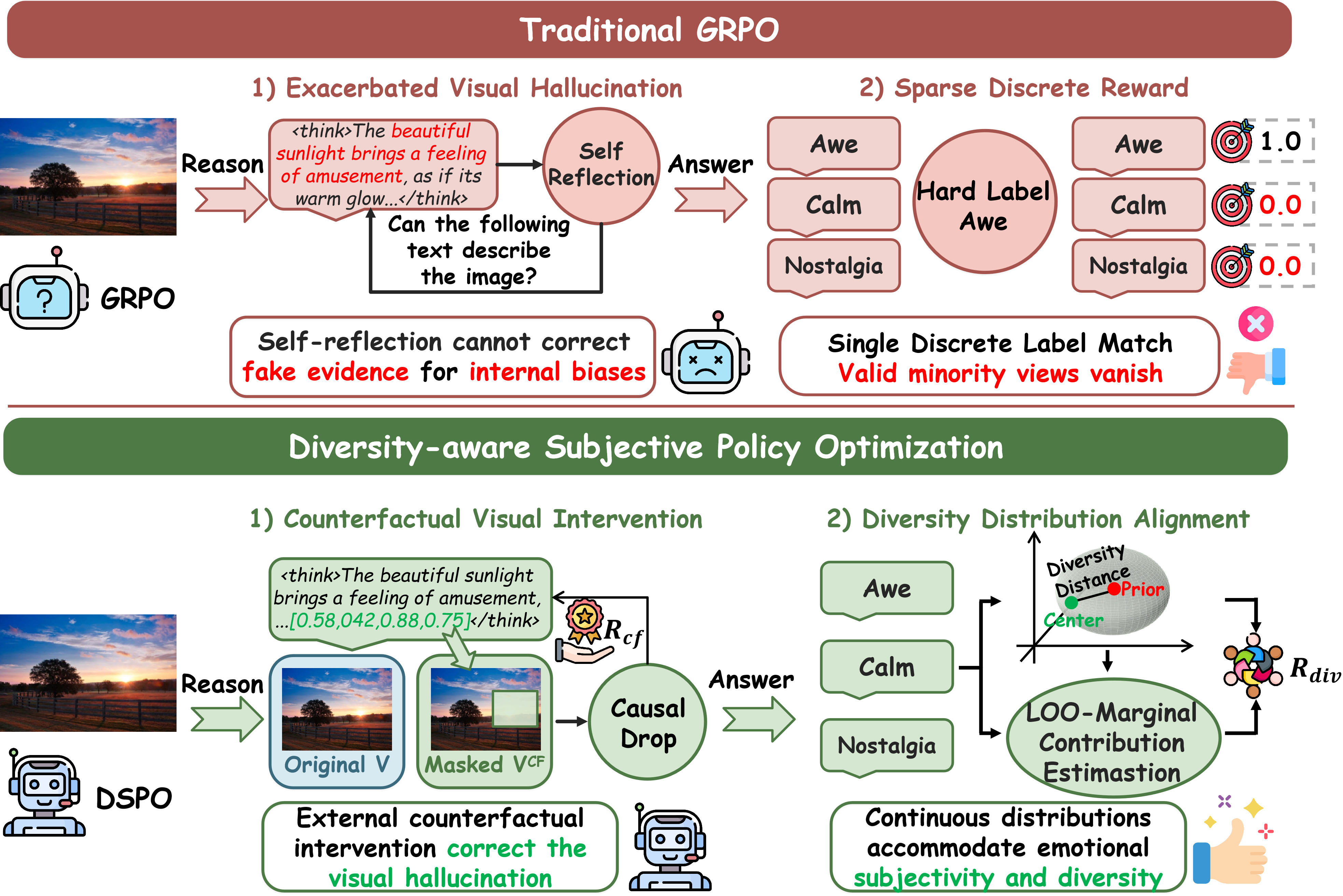}} 
\caption{Comparison between traditional GRPO and our DSPO. 1) Traditional self-reflection mechanisms struggle to correct the internal biases of MLLMs. Counterfactual visual intervention mitigates hallucinations about visual evidence. 2) Diversity Distribution Reward addresses the issue where the sparse, discrete rewards of GRPO struggle to accommodate sentiment-close answers.}
\label{fig1}
\vspace{-20pt}
\end{figure*}

To address these limitations, we propose Distribution-level Subjective Policy Optimization (DSPO), a novel reinforcement learning framework specifically tailored to align MLLMs with both human subjectivity and objective visual causality. DSPO constructs a context-grounded emotional distribution prior as an enhanced optimization signal by combining the lexical prior of the annotated emotion with image-specific contextual information. Then DSPO introduces a Distribution-Aligned Emotional Diversity Reward, which calculates the leave-one-out marginal contribution of each candidate emotion within a rollout and rewards candidates who makes the predicted entire distribution closer to the prior, thereby preserving plausible subjective interpretations without encouraging unconstrained dispersion.
Besides, to ensure this empirical distribution is not contaminated by hallucinated reasoning, we introduce a Counterfactual Visual Intervention Gating (CVIG). By enforcing explicit physical grounding and computing the causal probability drop under latent visual masking, CVIG strictly penalizes visual hallucinations and assigns a weight to each candidate emotion. Ultimately, DSPO harmonizes the subjective diversity of emotional expression with the causal validity of visual evidence, significantly enhancing the emotional reasoning capabilities of MLLMs.
In summary, our main contributions are as follows:

$\bullet$ We propose Diversity-aware Subjective Policy Optimization (DSPO), a novel RL framework that introduces a continuous emotion distribution prior as an enhanced optimization signal, addressing the sparse reward by discrete hard-labels in existing RL training for emotion reasoning tasks.

$\bullet$ We introduce a Distribution-Aligned Emotional Diversity Reward that calculates the subjective diversity of all generated candidate emotions within a rollout, naturally tolerating human emotional subjectivity without mode collapse. Besides, we design a Counterfactual Visual Intervention Gating, which verifies the validity of visual evidence by evaluating drops of causal probability under visual masking, alleviating exacerbated visual hallucinations in implicit affective reasoning.

$\bullet$ Extensive experiments demonstrate that DSPO achieves state-of-the-art performance on multiple emotion reasoning benchmarks, especially on the cross-domain performance, \emph{i.e.,} improving +10.8\% on average cross-domain accuracy than EMO-R3.
\begin{figure*}[t]
\center{\includegraphics[width=0.98\linewidth] {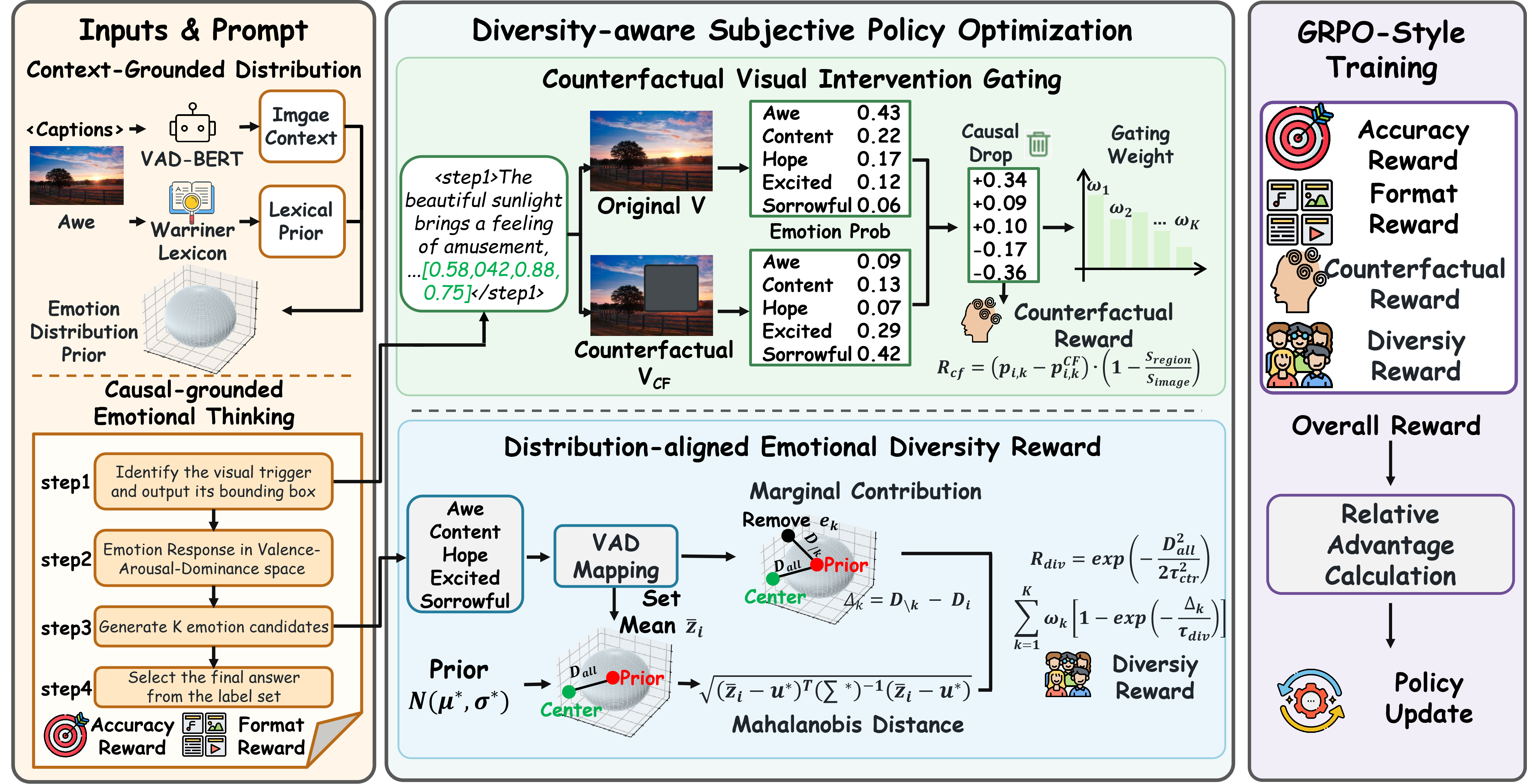}}
\caption{Overview of DSPO framework. The left part presents the construction of the emotional distribution prior and the input prompt.
The middle part show two crucial counterfactual visual intervention gating and distribution-aligned emotional diversity reward modules. Finally, DSPO are jointly optimized with the original Format and Accuracy rewards under the GRPO framework.}
\label{fig2}
\end{figure*}
\section{DSPO: Diversity-aware Subjective Policy Optimization}

\subsection{Preliminary}\label{3.1}
We evaluate the emotional reasoning ability using an image emotion recognition task. It is crucial to design an explicitly guided instruction prompt for the internal thinking phase of MLLMs. Generic Chain-of-Thought (CoT) prompts (\emph{e.g.}, "Please think step by step.") inherently lack task-specific designs. Consequently, they fail to establish a causal mapping between low-level visual cues and high-level abstract emotional conclusions. Such unconstrained exploration struggles to elicit reasoning trajectories that align with human affective cognition, and instead frequently exacerbates factual drift and visual hallucinations. To overcome this, we design a Cause-grounded Emotional Thinking.\footnote{The details of the thinking template are shown in the Appendix.} Specifically, we first guide the MLLM identify key visual evidence and output its location by a bounding box. Based on this visual evidence, we require the MLLM to output emotional responses. Subsequently, unlike previous methods, we do not directly require the MLLM to output a single emotion category. Instead, we allow the MLLM to first generate $K$ potential emotion candidates and then combine them to select the best match from the given label set. Such thinking mode enhances the mining of visual evidence through physical anchoring, while increasing the diversity of emotion reasoning by generating a candidate set of emotions. Overall, we adopt a GRPO-style framework. Given an image-prompt input pair $(\mathcal{V},{Q})$, the MLLM $\pi_{\theta_{old}}$ receives the input pair and samples a group of $G$ distinct rollouts, denoted as $\mathcal{O} = \{o_1,o_2,…,o_G\}$. Finally, the MLLM is optimized by maximizing the following objective function:
\begin{align}
 \mathcal{J}(\theta) =& \mathbb{E}_{(\mathcal{V},{Q})}E_{\mathcal{O} \sim \pi_{\theta_{old}}} \left[ \frac{1}{G} \sum_{i=1}^G \Big( \min \left[ \rho_{i}(\theta)  \hat{A}_i, \, \text{clip}\left(\rho_{i}(\theta) , 1-\epsilon, 1+\epsilon\right) \hat{A}_i \right] \right]- \beta\mathbb{D}_{KL}(\pi_\theta \,||\, \pi_{ref}) 
\end{align}
where $\epsilon,\beta$ are clipping and divergence penalty hyper-parameters and the importance sampling ratio $\rho_{i}(\theta)$ is defined as $\rho_{i}(\theta) = \frac{\pi_\theta(o_{i} \mid (\mathcal{V},{Q}))}{\pi_{\theta_{old}}(o_{i} \mid (\mathcal{V},{Q}))}$.

\subsection{Context-Grounded Emotional Distribution Prior}\label{3.2}
To provide a reliable prior for continuous emotion distributions. We design a context-grounded emotional distribution prior construction pipeline. A natural initial intuition is to leverage the widely used Warriner lexicon~\citep{warriner2013norms} to map discrete emotion labels into continuous VAD-Gaussian distributions. However, this approach completely ignores the nuanced affective variations induced by diverse visual scenes in the real world.
Therefore, we further introduce a dynamic context driven by the specific image content. Specifically, for the lexical prior, we query the Warriner lexicon to retrieve the baseline distribution parameters representing human consensus. For the dynamic context, we first process the image through an advanced MLLM Mimo-v2~\citep{xiao2026mimo} to generate a detailed caption. This caption is subsequently fed into a pre-trained sentence-level VAD regression model~\citep{buechel2017emobank} to extract context-specific distribution parameters:
\begin{equation}
    \boldsymbol{\mu}_{sta}, \boldsymbol{\sigma}_{sta} = \Phi_{lex}(y), \mathcal{C} = {\rm CapGen}(\mathcal{V}),\boldsymbol{\mu}_{ctx} = {\rm VADReg}(\mathcal{C}),
\end{equation}
where $y$ is the ground-truth emotion category and $\Phi_{lex},{\rm CapGen},{\rm VADReg}$ denote Warriner lexicon, caption generator, and VAD regression model, respectively. Besides, to quantify the ambiguity between the image context and lexical prior, we calculate the contextual divergence vectors:
\begin{equation}
    \boldsymbol{\sigma}_{ctx} = {\rm Var}(\boldsymbol{\mu}_{sta} , \boldsymbol{\mu}_{ctx}),
\end{equation}
where ${\rm Var}$ denotes the variance calculation. Finally, we derive the ultimate ground-truth distribution label by aggregating these two sets of parameters via weighted fusion:
\begin{equation}
    \boldsymbol{\mu}^* =  \frac{\boldsymbol{\mu}_{sta} +  \boldsymbol{\mu}_{ctx}}{2},\boldsymbol{\sigma}^* = \frac{\boldsymbol{\sigma}_{sta} +\boldsymbol{\sigma}_{ctx}}{2},
\end{equation}
The Gaussian distribution ${\mathcal{Z}} \sim \mathcal{N}(\boldsymbol{\mu}^*, \boldsymbol{\sigma}^*)$ represents the ground-truth human emotion distribution, serving as the gold standard for calculating emotion diversity.

\subsection{Counterfactual Visual Intervention Gating}
Prior works~\citep{fang2026emo} attempt to mitigate hallucinations through reflective emotional rewards. However, these approaches typically restrict themselves to prompt-level self-correction over existing outputs, leaving the internal self-bias of MLLMs fundamentally unaddressed. Moreover, this pattern encourages the MLLM to generate fake emotional cues to justify a correct emotion conclusion. We analyze that the core question is whether the generated textual claims actually point to a visually present region. Such causal attribution regarding the reasoning chain is independent of external emotional labels. To this end, we introduce a Counterfactual Visual Intervention Gating (CVIG) module. Concretely, we generate a counterfactual image by masking the pointed visual regions and measure the shift in prediction probabilities for each candidate emotion, which quantifies the consistency between the specified visual region and each emotional interpretation.
Specifically, we first extract the emotional cue text and its associated bounding box coordinates from the output:
\[
(t_i^1,[x_i^{min},y_i^{min},x_i^{max},y_i^{max}]) = \Phi_{regex}\bigl(o_i, \verb|<step1>|(.*?)\verb|</step1>|\bigr),
\tag{5}
\]
where $t_i^1$ represents the text segment for the visual evidence and $\Phi_{regex}$ denotes the regex extractor. Subsequently, we map the extracted bounding box coordinates to the indices of the visual encoder:
\[
\hat{x}_i^{\min} = \lfloor x_i^{\min} \cdot W \rfloor,\quad
\hat{x}_i^{\max} = \lceil x_i^{\max} \cdot W \rceil,\quad
\hat{y}_i^{\min} = \lfloor y_i^{\min} \cdot H \rfloor,\quad
\hat{y}_i^{\max} = \lceil y_i^{\max} \cdot H \rceil,
\tag{6}
\]
\[
\mathcal{T}_i = \left\{ (r, c) \mid \hat{y}_i^{\min} \leq r < \hat{y}_i^{\max},\; \hat{x}_i^{\min} \leq c < \hat{x}_i^{\max} \right\},
\tag{7}
\]
where $H,W$ is the size of the image. For visual features $\mathcal{V}_i\in\mathbb{R}^{H\times W\times D}$, we obtain the counterfactual image by constructing a mask matrix $M_i\in\mathbb{R}^{H\times W}$:
\[
M_i(j,k) = 
\begin{cases} 
0, & (j,k)\in\mathcal{T}_i,\\
1, & \text{otherwise},
\end{cases}
\quad\text{and}\quad
\mathcal{V}^{CF}_i = \mathcal{V}_i \odot M_i,
\tag{8}
\]
where $\odot$ denotes element-wise multiplication. We feed the original and counterfactual image features $\mathcal{V}_i$ and $\mathcal{V}^{CF}_i$ into the subsequent decoder and calculate the predicted probabilities for each of the $K$ candidate emotion categories extracted from the \verb|<step3>| list:
\[
p_{i,k} = P\bigl(e_i^{(k)} \mid \mathcal{V}_i, \mathcal{Q}, t_i\bigr),\qquad
p_{i,k}^{CF} = P\bigl(e_i^{(k)} \mid \mathcal{V}_i^{CF}, \mathcal{Q}, t_i\bigr),
\quad k=1,2,\ldots,K,
\tag{9}
\]
where $e_i^{(k)}$ denotes the $k$-th emotion candidate from the \verb|<step3>| list. We then compute the causal drop for each candidate as the relative decline in its prediction probability under masking, with an area penalty to prevent the model from selecting the entire image. Finally, we compute the total drop of all candidates as the counterfactual reward $\mathcal{R}_{cf}$:
\[
\Delta P_{i,k} = \bigl(p_{i,k} - p_{i,k}^{CF}\bigr) \cdot \left(1 - \frac{S_{\text{region}}}{S_{\text{image}}}\right),\quad\mathcal{R}_{cf,i}=\frac{1}{K}\sum_{k=1}^K\Delta P_{i,k}
\tag{10}
\]
where $S_{\text{region}}$ and $S_{\text{image}}$ denote the area of the anchored region and the entire image, respectively. Besides, we compute the normalized counterfactual weights across all $K$ candidates:
\[
w_{i,k} = \frac{\exp(\Delta P_{i,k} / \tau)}{\sum_{j=1}^{K}\exp(\Delta P_{i,j} / \tau)},\qquad \tau > 0,
\tag{11}
\]
where $\tau$ is the temperature parameter controlling the sharpness of the distribution. These gating weights are directly applied to the diversity reward to down-weight hallucinated candidates.

\subsection{Distribution-aligned Emotional Diversity Reward}\label{3.3}
To address the limitation of single discrete label and align the continuous and subjective human distribution, we introduce a distribution-aligned emotional diversity reward that encourages the model to produce a distribution of plausible emotional interpretations based on the constructed prior.

Specifically, we first extract the text segment of emotion response and $K$ candidate emotions within each rollout:
\[
t_i^2,\left\{e_i^{(1)}, e_i^{(2)}, \ldots, e_i^{(K)}\right\} = \Phi_{regex}\bigl(o_i, [\verb|<step2>|(.*?)\verb|</step2>|,\verb|<step3>|(.*?)\verb|</step3>|]\bigr),
\tag{12}
\]
then each emotion word \(e_i^{(k)}\) is mapped to a VAD vector using the Warriner lexicon:
\[
\mathbf{z}_i^{(k)} = \Phi_{\text{lex}}\bigl(e_i^{(k)}\bigr) \in \mathbb{R}^3, \quad k = 1, 2, \ldots, K,
\tag{13}
\]
where the three dimensions correspond to Valence, Arousal, and Dominance, respectively. For out-of-vocabulary words, we apply a stemmer-based fallback or nearest-neighbor lookup to ensure robust coverage.
Subsequently, to capture the subjective diversity of emotional interpretations beyond mere dispersion, we adopt a Leave-One-Out (LOO) marginal contribution estimation. We first compute the VAD-vector centroid of $K$ candidate emotions. The diversity baseline is defined as the Mahalanobis distance from the centroid to our constructed prior distribution:
\[
\bar{\mathbf{z}}_i = \frac{1}{K} \sum_{k=1}^{K} \mathbf{z}_{i}^{(k)},\quad
D_{i,all} = \sqrt{(\bar{\mathbf{z}}_i - \boldsymbol{\mu}^*)^\top (\boldsymbol{\Sigma}^*)^{-1} (\bar{\mathbf{z}}_i - \boldsymbol{\mu}^*)},
\tag{14}
\]
where $\boldsymbol{\Sigma}^*=\begin{vmatrix}
\sigma_V & &   \\
 & \sigma_A &   \\
  &  &\sigma_D \\
\end{vmatrix}$ is the covariance matrix. Subsequently, for each candidate, we remove it and recompute the distance using the remaining \(K-1\) candidates:
\[
D_{i,\setminus k} = \sqrt{(\bar{\mathbf{z}}_{i,\setminus k} - \boldsymbol{\mu}^*)^\top (\boldsymbol{\Sigma}^*)^{-1} (\bar{\mathbf{z}}_{i,\setminus k} - \boldsymbol{\mu}^*)},
\quad \bar{\mathbf{z}}_{i,\setminus k} = \frac{1}{K-1} \sum_{j \neq k} \mathbf{z}_{i}^{(j)},
\tag{15}
\]
the marginal contribution of $e_i^{(k)}$ is defined as the increase in distance. A positive increase indicates that the candidate carries a unique subjective perspective that brings the group closer to the human prior. Then we design the diversity reward $\mathcal{R}_{div}$ by considering the centroid distance and the sum of the marginal contribution of all candidates:
\[
\Delta_{i,k} = D_{i,\setminus k} - D_{i},\quad
\mathcal{R}_{div,i} = \exp\left(
-\frac{D_{i,all}^2}{2\tau_{\mathrm{ctr}}^2}
\right) \sum_{k=1}^{K} w_{i,k}\left[
1-\exp\left(
-\frac{\Delta_{i,k}}{\tau_{\mathrm{div}}}
\right)
\right],
\tag{16}
\]
where $\tau_{\mathrm{ctr}},\tau_{\mathrm{div}}$ are temperature coefficient. By comprehensively considering the accuracy of the central distribution and the internal emotional diversity of the candidate set, we guide the MLLM to achieve interpretable emotional reasoning that aligns with human subjective diversity.

\subsection{Overall Reward and Training}\label{3.4}
Besides the above two designed rewards $\mathcal{R}_{cf}$ and $\mathcal{R}_{div}$, following the traditional GRPO training, we define two general rewards to guide the optimization of the structured emotional reasoning. We first define the format reward $\mathcal{R}_{fmt}$ to measure whether the generated reasoning text adheres to the pre-defined structure. Specifically, it checks whether each reasoning step corresponds to the expected stage \verb|<stepi>…</stepi>| and whether the bounding box, the emotion list, and the chosen option are correctly enclosed in \verb|\bboxed{}|, \verb|\list{}|, and \verb|\boxed{}|:
\[
    \mathcal{R}_{\text{format}} = \begin{cases} 
1, & \text{if}\quad\verb|\bboxed{}|, \verb|\list{}|, \verb|\boxed{}|\quad\text{formats are correct,}\\ 
0, & \text{otherwise.} 
\end{cases}
\tag{17}
\]
Meanwhile, the accuracy reward $\mathcal{R}_{acc}$ evaluates whether the chosen option $\hat{\mathcal{E}}$ aligns with the ground-truth emotion label $\mathcal{E}^*$:
\[
    \mathcal{R}_{\text{acc}} = \begin{cases} 
1, & \text{if } \hat{\mathcal{E}} = \mathcal{E}^*, \\ 
0, & \text{otherwise.} 
\end{cases}
\tag{18}
\]
Finally, the total reward of the DSPO training is calculated by the weighted sum of all four rewards:
\[
\mathcal{R}_{total} = \lambda_{acc} \cdot \mathcal{R}_{acc}
+ \lambda_{fmt} \cdot \mathcal{R}_{fmt}
+ \lambda_{cf} \cdot \mathcal{R}_{cf}
+ \lambda_{{div}} \cdot \mathcal{R}_{div},
\tag{19}
\]
where $\lambda_{acc}, \lambda_{fmt},\lambda_{cf},\lambda_{div}$ are hyper-parameters to control the balance between different rewards.

\begin{table}[t]
\centering
\caption{\label{main}{Comparison with GRPO variants and SOTA methods across in-domain and out-of-domain settings. The best and suboptimal results are highlighted in bold and underline, respectively.}}
\scalebox{0.75}{
\begin{tabular}{c|c|c|cc|c|cc|ccc}
\toprule
{{Methods}}  &  {{Rollout}} & {{EmoSet$^I$}} & {{Emotion6}} & {{WebEmo}} & {{Emotion6$^I$}} & {{EmoSet}} & {{WebEmo}} & {$\mathcal{A}^I$} &  {$\mathcal{A}^O$} &  {$\mathcal{A}$} \\
\toprule
\multicolumn{11}{l}{{\textcolor{gray!50}{\textit{LLaVA-1.5-7B}}}}\\
\rowcolor{gray!15}{Zero-shot}  & - & 52.77& 48.32& 25.56 &48.32& 52.77 &25.56& 50.55& 38.05& 42.22\\
SFT& -& 56.04& 54.21 &42.39& 54.21& 56.04& 42.39& 55.13 &48.76& 50.88\\
\cmidrule{1-11}
\multicolumn{11}{l}{{\textcolor{gray!50}{\textit{Qwen2.5-VL-3B-Instruct}}}}\\
\rowcolor{gray!15}{Zero-shot}  & - & 51.55 & 50.00 & 40.65 & 50.00 & 51.55 & 40.65 & 50.77 & 45.71 & 47.40\\
SFT& -& \cellcolor{oursburgundy}{77.15}& 34.51& 17.75& 69.53& 26.45& 37.65& \cellcolor{oursburgundy}{73.34}& 29.09& 43.84\\
\cmidrule{1-11}
\rowcolor{gray!15}GRPO~\citeyearpar{shao2024deepseekmath}& \multirow{4}{*}{4}& 74.60& 60.10& 49.50& \underline{70.88}& 59.90& 44.85& 72.74& 53.59& 59.97\\
DAPO~\citeyearpar{yu2026dapo}& ~& 68.99& 56.90& 49.80& 68.56& 59.95& \underline{45.50}& 68.78& 53.04& 58.28\\
EMO-R3~\citeyearpar{fang2026emo}& ~& \textbf{75.50}& \underline{60.44}& \underline{50.45}& 70.71& \underline{60.70}& 45.20 &\underline{73.10}& \underline{54.20} &\underline{60.50}\\
\rowcolor{oursburgundy}\textbf{Ours}&~&\underline{75.30} &\textbf{67.31} &\textbf {54.54}	&\textbf{71.20}	&\textbf{69.80}	&\textbf{51.80}	&\textbf{73.25} &\textbf{60.86}	&\textbf{64.99} \\
\cmidrule{1-11}
\rowcolor{gray!15}GRPO~\citeyearpar{shao2024deepseekmath}& \multirow{4}{*}{8}& 75.45& 57.91 &49.40 &69.87 &60.30 &42.05& 72.66 &52.42& 59.16\\
DAPO~\citeyearpar{yu2026dapo} &~&70.21& 55.72& 48.80 &62.39& 58.05 &\underline{46.30} &66.30 &52.22 &56.91\\
EMO-R3~\citeyearpar{fang2026emo}&~& \underline{76.40}& \underline{59.26}& \underline{49.70}&\underline{71.72}& \underline{61.80}& 43.65& \underline{74.06}& \underline{53.60}& \underline{60.42}\\
\rowcolor{oursburgundy}\textbf{Ours}&~&\textbf{76.65} &\textbf{66.87} &\textbf {54.00}	&\textbf{71.80}	&\textbf{67.40}	&\textbf{49.25}	&\textbf{74.23} &\textbf{59.38}	&\textbf{64.33} \\
\bottomrule
\end{tabular}}
\vspace{-15pt}
\end{table}

\section{Result and Discussion}

\subsection{Main Results}
As shown in Table 1, we first observe that DSPO achieves the highest overall accuracy of in- and out-of- domain under both \(G=4/8\), \emph{i.e.,} +46.7\%/+8.7\% improvements than SFT/EMO-R3 when $G=8$. This demonstrates the superior performance of DSPO across various scenarios involving emotion understanding.
Besides, we observe that the improvements are particularly pronounced under out-of-domain evaluation, \emph{i.e.,} +13.6\%/+12.3\% improvements than GRPO/EMO-R3 when $G=8$. Such significant gains indicate that DSPO improves more than in-domain label fitting. The distribution-aligned diversity reward preserves plausible neighboring affective interpretations in the continuous VAD space, while CVIG suppresses candidates unsupported by visual evidence, jointly reducing label-specific shortcut learning. Meanwhile, DSPO maintains a competitive in-domain performance and the in-domain accuracy under two rollout settings is slightly higher than EMO-R3, which indicates that diversity reward learning does not come at the expense of fitting in-domain labels and instead enhances the affective semantic space upon that foundation.

\begin{wrapfigure}{r}{0.65\textwidth} 
  \centering
  \setlength{\aboverulesep}{0pt}
\setlength{\belowrulesep}{0pt}
  \vspace{-10pt} 
\caption*{Table 2: The results for distribution-based evaluation.}
\label{tab:loo}
\small
\begin{tabular}{c|cc|cc}
\toprule
{{Methods}} & {{KL$\downarrow$}} & {{JS$\downarrow$}} & {{$H^{model}\uparrow$}} & {{$\text{En-MAE}\downarrow$}} \\
\toprule
\rowcolor{gray!15}SFT   & 1.940  & 0.457   &  0.395  & 0.397  \\
GRPO    & 0.702  &  0.260  &  0.581  & 0.298  \\
GRPO+$\mathcal{R}_{en}$& 0.884  &  0.295  &  \textbf{0.763}  & 0.329  \\
\rowcolor{gray!15}DAPO   & 1.059  &  0.328  &  0.524  & 0.348  \\
EMO-R3    & \underline{0.650}  & \underline{0.244}   & 0.623   &  \underline{0.274} \\
\rowcolor{oursburgundy} \textbf{DSPO} & \textbf{0.372} & \textbf{0.169} & \underline{0.696} & \textbf{0.209} \\
\bottomrule
\end{tabular}
\end{wrapfigure}

We also evaluate on four distribution-based metrics.\footnote{Settings of dataset, metrics, and implementation details are shown in the Appendix.} As shown in Table 2, we first observe that DSPO achieves the best performance on three relative metrics, \emph{i.e.,} KL, JS, and En-MAE. This indicates that DSPO could generate sampling distributions that are closer to the true human distribution. Furthermore, we observe that adding an entropy-based reward to GRPO improves $H^{model}$ but leads to a significant decline across three relative metrics. This demonstrates that the unconstrained diversity generation fails to enhance emotional reasoning capabilities. In contrast, DSPO calculates a diversity reward aligned with the human distribution, enabling the model to achieve a balance between accurate emotional reasoning and diverse emotional generation.

\subsection{Ablation Studies}

\begin{wrapfigure}{r}{0.65\textwidth} 
  \centering
  \setlength{\aboverulesep}{0pt}
\setlength{\belowrulesep}{0pt}
  \vspace{-10pt} 
\caption*{Table 3: The ablation study for CVIG and DEDR modules.}
\label{tab:ablation_module}
\begin{tabular}{cc|c|cc|c}
\toprule
CVIG & DEDR & EmoSet$^I$ &Emotion6 &WebEmo &$\mathcal{A}$ \\
\midrule
\rowcolor{gray!15}$\times$ & $\times$ &75.45 &57.91 &49.40 &60.92 \\
$\checkmark$ & $\times$ &76.10 &58.85 &48.91 &61.29 \\
$\times$ & $\checkmark$ &75.60 &64.25 &52.60 &64.15 \\
\rowcolor{oursburgundy}$\checkmark$ & $\checkmark$ & \textbf{76.65} & \textbf{66.87} & \textbf{54.00} & \textbf{65.84} \\
\bottomrule
\end{tabular}
  \vspace{-10pt} 
\end{wrapfigure}

\textbf{Discussion on proposed modules.} Table 3 explores the contributions of CVIG and DEDR. First, we observe that using CVIG alone primarily improves in-domain accuracy and even degrades on the out-of-domain WebEmo dataset. CVIG enhances the authenticity and accuracy of emotional reasoning by filtering out spurious visual evidence. Besides, using DEDR alone significantly improves out-of-domain performance. By designing a diversity reward that guides the MLLM to learn continuous emotion distributions rather than only fitting the label distribution of training source via hard labels, DEDR substantially enhances the robustness. Finally, the synergy between the CVIG and DEDR modules further enhances overall performance, thereby enabling a more comprehensive understanding of emotion.

\begin{wrapfigure}{r}{0.55\textwidth} 
  \centering
  \setlength{\aboverulesep}{0pt}
\setlength{\belowrulesep}{0pt}
  \vspace{-10pt} 
\caption*{Table 4: The ablation study for the number of candidate emotions.}
\label{tab:ablation_topk}
\begin{tabular}{c|c|cc|c}
\toprule
{{Top-$K$}} & {{EmoSet$^I$}} & {{Emotion6}} & {{WebEmo}} & {{$\mathcal{A}$}} \\
\toprule
\rowcolor{gray!15}1   & 72.35  &  59.20  &  50.15  & 60.57  \\
 3    & 75.90  &  62.74  &  51.20  & 63.28  \\
\rowcolor{oursburgundy} \textbf{5} & \textbf{76.65} & \textbf{66.87} & \textbf{54.00} & \textbf{65.84} \\
 7    & 74.45  &  60.81  &  50.99  &  62.08 \\
\bottomrule
\end{tabular}
\end{wrapfigure}

\textbf{Impact of Candidate Emotion Number.} Table 4 explore the impact of different numbers of candidate emotions within each rollout. We first observe a significant performance drop when $K=1$. A single candidate is difficult to adequately represent the ambiguity of human emotions and fit the distribution prior. A moderate candidate set allows the model to cover multiple plausible regions of the target VAD distribution and provides more informative marginal-contribution estimates. Nevertheless, model performance begins to decline as increasing $K=7$, suggesting that an excessively large candidate set introduces redundant grounded emotions and adds noise to distribution matching. Finally, we set \(K=5\) as an effective balance between subjective coverage and candidate reliability.

\begin{wrapfigure}{r}{0.65\textwidth} 
  \centering
  \setlength{\aboverulesep}{0pt}
\setlength{\belowrulesep}{0pt}
  \vspace{-10pt} 
\caption*{Table 5: The discussion for diversity computation.}
\label{tab:loo}
\small
\begin{tabular}{c|c|cc|c}
\toprule
{{Setting}} & {{EmoSet$^I$}} & {{Emotion6}} & {{WebEmo}} & {{$\mathcal{A}$}} \\
\toprule
\rowcolor{gray!15}w/o DEDR & 76.10 & 58.85 & 48.91 & 61.29 \\
\cmidrule{1-5}
Center Distance   &  76.30 &  60.01  &  48.75  & 61.69  \\
\rowcolor{gray!15}Pairwise Distance    & 74.74  &  58.40  &  47.89  & 60.34  \\
LOO-Margin Distance    & 75.09  &  65.49  & 52.60   & 64.39  \\
\rowcolor{oursburgundy} \textbf{Combined Distance} & \textbf{76.65} & \textbf{66.87} & \textbf{54.00} & \textbf{65.84} \\
\bottomrule
\end{tabular}
  \vspace{-10pt} 
\end{wrapfigure}

\textbf{Discussion on diversity computation.} Table 5 compares different diversity computation strategies, \emph{i.e.,} 1) Center Distance: Only calculate the distance between predicted entire distribution and prior. 2) Pairwise Distance: Calculate the distances between all pairs of candidate emotions. 3) LOO-Margin Distance: Only calculate the leave-one-out marginal contribution of each candidate emotion. We first observe that Pairwise Distance significantly degrades overall performance, indicating that focusing solely on in-set diversity without constraining it close to the prior is insufficient. Besides, considering center and LOO-Margin distance separately both fail to achieve the best performance, which indicates that we need to comprehensively consider both the accuracy of the entire set within the emotional semantic space and the diversity within the set.

\begin{wrapfigure}{r}{0.65\textwidth} 
  \centering
  \setlength{\aboverulesep}{0pt}
\setlength{\belowrulesep}{0pt}
  \vspace{-10pt} 
\caption*{Table 6: The discussion on CVIG module.}
\label{tab:cvig}
\small
\begin{tabular}{c|c|cc|c}
\toprule
{{Setting}} & {{EmoSet$^I$}} & {{Emotion6}} & {{WebEmo}} & {{$\mathcal{A}$}} \\
\toprule
\rowcolor{gray!15}w/o CVIG   & 75.60  & 64.25   & 52.60   &  64.15 \\
\cmidrule{1-5}
\multicolumn{5}{l}{{\textcolor{gray!50}{\textit{Bounding Box}}}}\\
\rowcolor{gray!15}Random Box    & 67.80  &  57.30  &  46.59  &  57.23 \\
w/o Area Penalty    & 71.55  &  60.89  & 49.10   & 60.51  \\
\cmidrule{1-5}
\multicolumn{5}{l}{{\textcolor{gray!50}{\textit{Intervention}}}}\\
\rowcolor{gray!15}Mean Replace    & 72.11  & 60.60   &  49.70  & 60.80  \\
Gaussian Noise    & 75.29  &  64.50  &  52.89  & 64.23  \\
\cmidrule{1-5}
\rowcolor{oursburgundy} \textbf{w/ CVIG} & \textbf{76.65} & \textbf{66.87} & \textbf{54.00} & \textbf{65.84} \\
\bottomrule
\end{tabular}
  \vspace{-10pt} 
\end{wrapfigure}

\textbf{Discussion on CVIG module.} Table 6 explores the effects of the bounding box and intervention settings in CVIG module. We first observe that using random box causes a substantial performance drop. This is due to the removal of emotion-related regions. Besides, removing the area penalty also degrades the average accuracy. Without this penalty, the model may prefer excessively large bounding boxes containing both relevant and irrelevant content, resulting in an imprecise counterfactual intervention.
We further compare different intervention strategies for the selected regions. We observe that both gaussian noise and mean replacement reduce the performance. Gaussian noise introduces additional visual redundancy, and mean replacement does not completely remove the semantic information of the selected region. In contrast, zero masking provides a cleaner intervention by explicitly suppressing the selected visual features, thereby producing a clearer difference between the original and counterfactual predictions.


\begin{wrapfigure}{r}{0.5\textwidth} 
     \centering
     \includegraphics[width=0.5\textwidth]{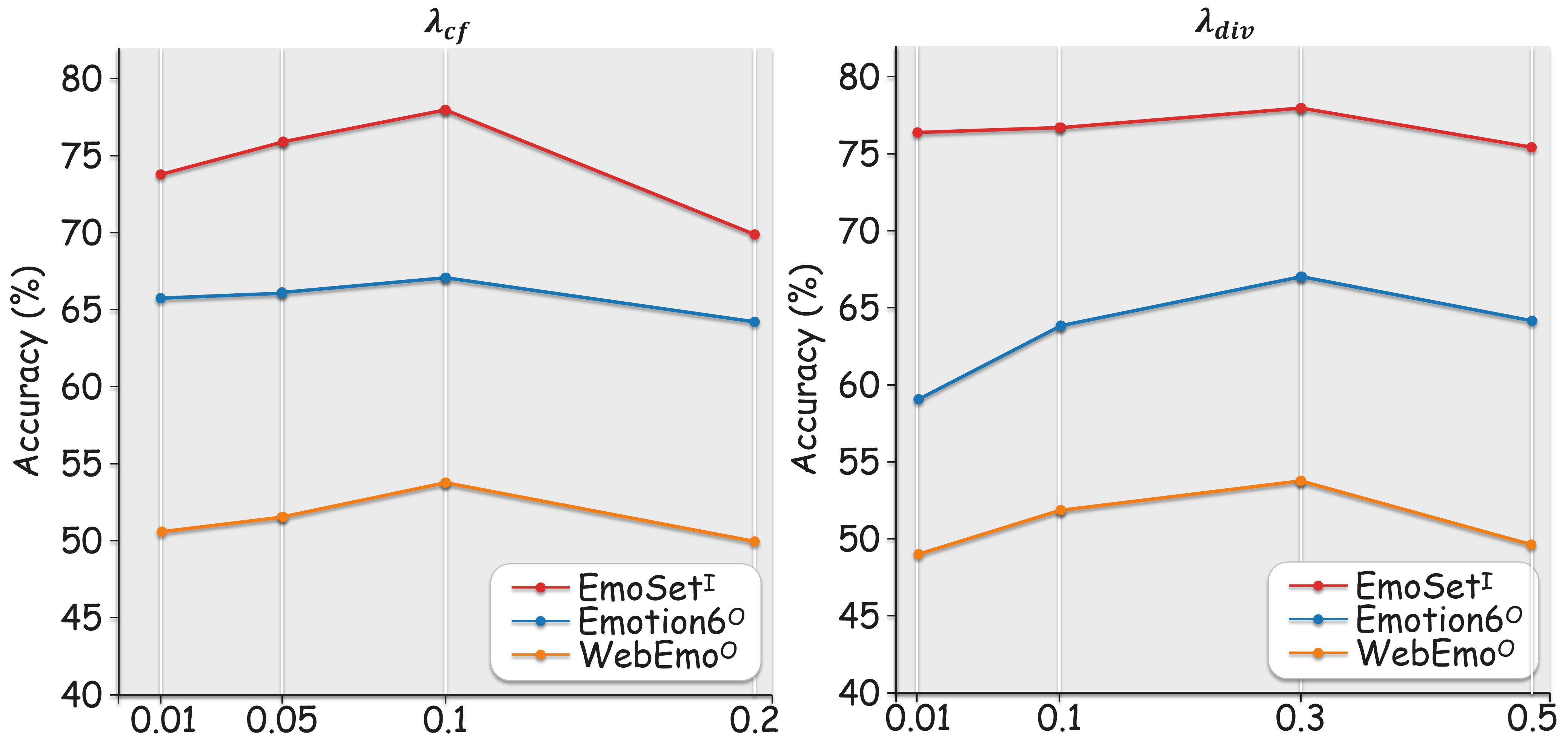} 
     \caption*{Figure {3}: Impact for different reward weights.}
     \label{param}
     \vspace{-15pt}
 \end{wrapfigure}

\textbf{Impact of Reward Hyper-parameters.} Fig. 3 investigates the sensitivity to $\lambda_{cf}$ and $\lambda_{div}$. We first observe that increasing $\lambda_{cf}$ from \(0.01\) to \(0.1\) mainly improves the in-domain performance. $\mathcal{R}_{cf}$ provides a grounding signal that suppresses fabricated emotional evidence. However, increasing \(\lambda_{cf}\) to \(0.2\) leads to a clear performance degradation. Overemphasizing visual consistency may amplify localization noise, suppress valid but subtle emotional cues.
Besides, increasing \(\lambda_{div}\) from \(0.01\) to \(0.3\) mainly brings a significant improvement on out-of-domain performance. By encouraging the model to cover multiple plausible emotional interpretations around the human affective distribution, $\mathcal{R}_{div}$ reduces over-reliance on a single hard label and improves the robustness. However, the performance decreases on both in-domain and out-of-domain when increasing \(\lambda_{div}\) to \(0.5\), which indicates that overemphasizing emotional diversity will affect basic emotional reasoning abilities.

\subsection{Efficiency Analysis} 
Considering that we introduce additional modules, we conduct an efficiency analysis on the training process under the 8-rollout setting on EmoSet dataset. As shown in Fig. 5, we observe that although our model introduces a certain amount of extra computation overhead, it does not bring about a significant improvement in training time. Compared with EMO-R3~\citep{fang2026emo}, our model

\begin{wrapfigure}{r}{0.5\textwidth} 
     \centering
     \vspace{-15pt}
     \includegraphics[width=0.5\textwidth]{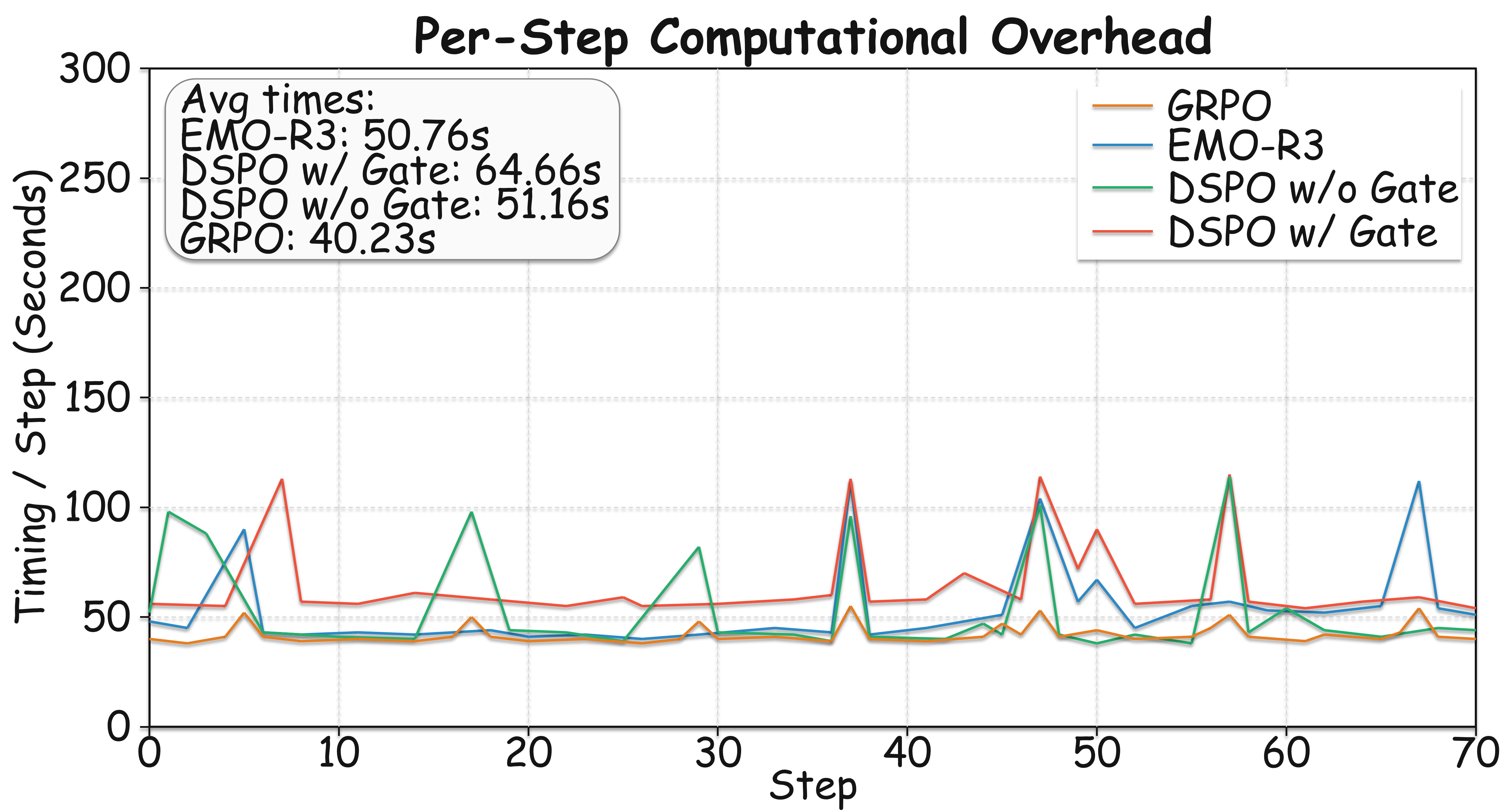} 
     \caption*{Figure {5}: Efficiency analysis visualization.}
     \label{time}
     \vspace{-20pt}
 \end{wrapfigure}

improves out-of-domain accuracy on the Emotion6 dataset by 12.8\% while consuming only 27\% more time. Moreover, the proposed gating and diversity module are both removed during the inference process, so that it requires no additional inference-time cost. Therefore, in real usage and evaluation, our model achieves better performance while maintaining high computational efficiency.

\subsection{Case Study}
We present a case study to compare between baseline EMO-R3~\citep{fang2026emo} and DSPO. As shown in Fig. 4, we first observe that EMO-R3 incorrectly identifies the image as `disgust', while DSPO correctly identifies it as `awe'. Furthermore, we analyze that the reason is that EMO-R3 erroneously localizes visual evidence as dark clouds and hallucinates a `heavy, oppressive environment'. Instead, from the weight distribution from CVIG, we find that DSPO detects this hallucination through counterfactual intervention and assigns the lowest weight to `sadness'. Besides, DSPO could estimate human-aligned diversity. Specifically, removing `sadness' reduces the overall distance, whereas removing `awe' increases it. Overall, DSPO learns within a continuous emotion space rather than simply fitting the label distribution of the dataset as previous methods do.
\begin{figure*}[t]        
\center{\includegraphics[width=0.95\linewidth] {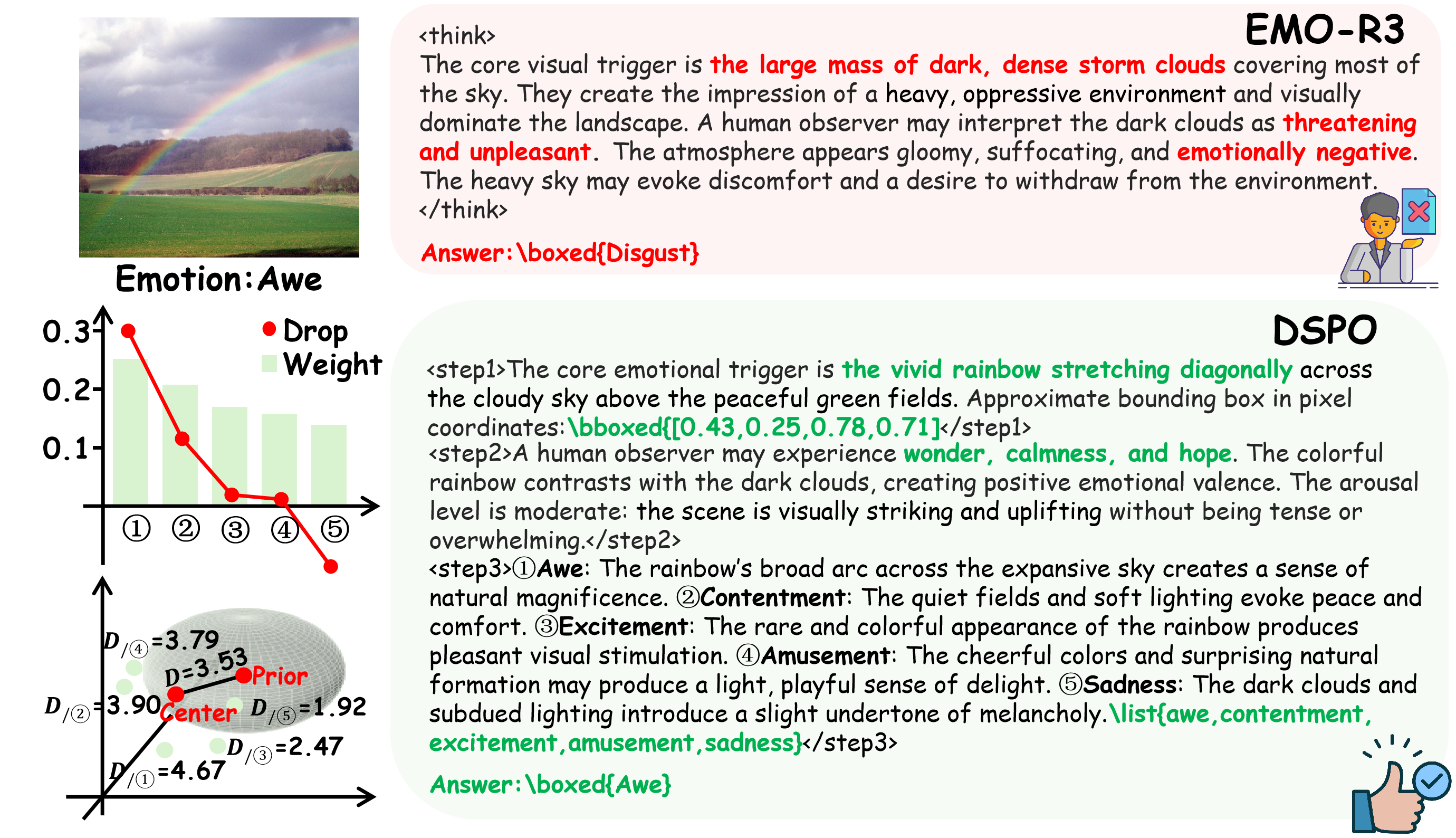}} 
\caption*{Figure {4}: Case study between the most powerful method EMO-R3 and DSPO on the EmoSet dataset.}
\label{cs}
\end{figure*}

\section{conclusion}
In this paper, we introduce Diversity-Aware Subjective Policy Optimization (DSPO) for robust emotion reasoning, which addresses two limitations of conventional RL learning: the sparse discrete reward and exacerbated visual hallucinations. Specifically, to achieve continuous emotion supervision, we first construct a context-grounded emotional distribution prior and propose a Distribution-Aligned Emotional Diversity Reward that evaluates the marginal contribution of each candidate emotion to align the prior. Besides, to alleviate hallucinations when searching for visual evidence, we further introduce Counterfactual Visual Intervention Gating, which estimates candidate-wise causal drop through counterfactual masking. Extensive experiments demonstrate that DSPO consistently improves overall accuracy and delivers particularly strong cross-domain generalization.

\bibliography{iclr2026_conference}
\bibliographystyle{iclr2026_conference}

\clearpage

\appendix
\section{related work}
\subsection{Multimodal Emotion Reasoning}
Multimodal emotion reasoning (MER) aims to infer emotional states and human intentions from multimodal information, including textual language, behavioral actions, speech signals, and social context. Initially, the community treats MER merely as a simple classification task. Some researchers develop modality fusion methods to enhance the ability to predict emotion categories~\citep{yang2023context,cheng2023semi,chen2026creatiparser,chen2026subjective}. However, given the complexity of human emotions, such simple fitting to ground truth lacks emotional interpretability, making it difficult for models to acquire genuine emotional reasoning capabilities. Recently, researchers shift the focus toward open-ended and interpretable emotional reasoning. AffectGPT~\citep{lian2025affectgpt} constructs a descriptive emotion dataset EMER-Coarse with 2K fine-grained emotion categories and designs a two-stage training framework to better align with manually-checked results. OV-MER~\citep{lian2025ov} proposes a novel paradigm to enable emotion prediction without being confined to predefined spaces and presents a newly curated database, novel evaluation metrics, and a preliminary benchmark. EmotionLLaMA~\citep{cheng2024emotion} integrates multimodal inputs and aligns multimodal features with instruction tuning to enhance the emotion reasoning. Furthermore, some studies have attempted to leverage reinforcement learning algorithms to bolster emotional reasoning abilities. 
R1-Omni~\citep{zhao2025r1} presents the first application of RL to an Omni-multimodal LLM for MER task, significantly enhancing the reasoning and generalization ability.
OmniOPSD~\citep{cheng2026omniopsd} utilizes the generated rationale a s privileged evidence accessible only to the teacher model, providing dense token-level scoring and supervision for the self-generated trajectories of student model.
EMO-R3~\citep{fang2026emo} proposes a reflective reinforcement learning framework, which leverages structured emotional thinking and reflective emotional reward to guide the model to perform emotion reasoning in an interpretable and step-by-step manner.
Despite overcoming the limitations of closed-set prediction, these studies still treat MER as a label-level task, overlooking the subjective nature and continuous distribution of emotions. Semantically similar emotions often coexist. Reliance on discrete emotion labels prevents models from learning to reason about emotions in a continuous manner. To address this limitation, DSPO focuses on group-level soft distributional matching by aggregating all generated rollouts into a joint empirical distribution, encouraging valid emotional diversity and enhancing continuous emotional reasoning.

\subsection{Hallucination Mitigation in MLLMs}
As the generative capabilities of MLLMs advance, the issue of multimodal hallucinations has become increasingly pronounced, which refers to the inconsistency between the generated text and the provided images~\citep{ye2025improving,song2025towards,hong2026emostyle,chen2023weakly}. This phenomenon may stem from an over-reliance on language priors, erroneous visual perception, or inadequate cross-modal reasoning~\citep{jain2024vcoder,li2024monkey,wang2023evaluation}. Researchers have explored various strategies to mitigate these hallucinations. Fine-tuning approaches focus on constructing high-quality datasets for fine-grained alignment to bridge the gap between visual and textual knowledge~\citep{you2024ferret,liu2024mitigating}. However, this demands valuable annotation costs and substantial computational resources. Alternatively, post-hoc methods utilize external tools or self-reflection mechanisms to correct hallucinated outputs~\citep{zhou2024analyzing,huang2024opera}. Moreover, certain decoding strategies delve into detecting anomalous attention tokens during generation, applying targeted interventions based on these observed patterns~\citep{tang2025intervening,wang2025mllm}.

Crucially, for multimodal emotion reasoning tasks, the exacerbation of multimodal hallucinations is remarkably severe. We infer this is because emotional cues are implicitly nested within abstract semantics, such as subtle micro-expressions, lighting, or overall atmospheric nuances, rather than explicit physical entities~\citep{ye2024dual,song2024emotional,ye2025multi}. Without specifically fine-tuning, existing MLLMs lack the intrinsic capability to mine these implicit emotion cues, leading to the frequent fabrication of visual facts to cater to emotional conclusions. To overcome this critical bottleneck, we introduce a Counterfactual Visual Intervention Gating (CVIG). By masking specific visual regions, CVIG generates a counterfactual image and computes the causal discrepancy in emotion prediction probabilities to evaluate the causal impact of the proposed visual cues. By rewarding genuinely causal visual evidence and penalizing hallucinated fabrications, CVIG effectively mitigates the multimodal hallucinations in emotion reasoning.

\subsection{Proof: DSPO is a Variational Lower Bound of Genuine Human Emotion}
In this section, we make a theoretical analysis~\citep{shannon1948mathematical} to prove that our proposed DSPO is a variational lower bound of genuine human emotion from an information-theoretic perspective. We follow the notations above: the image input $\mathcal{V}$, the output of MLLM $\mathcal{O}$, and the ground-truth human emotional distribution $\mathcal{Z}$, respectively. Overall, regarding our optimization objective:
\[
 \mathcal{J}(\theta) = \mathbb{E}_{(\mathcal{V},{Q})}E_{\mathcal{O} \sim \pi_{\theta_{old}}} \left[ \frac{1}{G} \sum_{i=1}^G \Big( \min \left[ \rho_{i}(\theta)  \hat{A}_i, \, \text{clip}\left(\rho_{i}(\theta) , 1-\epsilon, 1+\epsilon\right) \hat{A}_i \right] \right]- \beta\mathbb{D}_{KL}(\pi_\theta \,||\, \pi_{ref}) 
\tag{1}
\]
we aim for the text generated by the MLLM to exhibit the highest similarity with the true human emotion distribution given the visual prior, which is equivalent to maximizing the conditional mutual information $I(\mathcal{O}; \mathcal{Z} | \mathcal{V})$. Mathematically, it can be decomposed into the following form:
\[
I(\mathcal{O}; \mathcal{Z} | \mathcal{V}) = H(\mathcal{Z} | \mathcal{V}) - H(\mathcal{Z} | \mathcal{O}, \mathcal{V}),
\tag{2}
\]
$H(\mathcal{Z}|\mathcal{V})$ denotes the inherent uncertainty of human emotion given the image cues, which is a constant determined by human priors. Thus, our goal is to minimize $H(\mathcal{Z} | \mathcal{O}, \mathcal{V})$, which represents the residual uncertainty of human emotion, given the provided image cues and the response of MLLMs. Based on the information-theoretic definition, it is equivalent to maximizing the following expectation:
\[
-H(\mathcal{Z} | \mathcal{O}, \mathcal{V}) = \mathbb{E}_{\mathcal{O}, \mathcal{Z}} [\log P_{true}(\mathcal{Z} | \mathcal{O}, \mathcal{V})],
\tag{3}
\]
for $P_{true}(\mathcal{Z} | \mathcal{O}, \mathcal{V})$, it is an internal representation that is difficult to observe directly. Thus, we introduce Evidence Lower Bound (ELBO)~\citep{kingma2013auto} to approximate it.
\begin{align*}
    \mathbb{E}_{\mathcal{O}, \mathcal{Z}} [\log P_{true}(\mathcal{Z} | \mathcal{O}, \mathcal{V})] 
    &\ge \underbrace{ \mathbb{E}_{\mathcal{O} \sim \pi_\theta} \left[ \mathbb{E}_{\mathcal{Z} \sim P^*(\mathcal{Z}|\mathcal{V})} [\log Q(\mathcal{Z} | \mathcal{O})] \right] }_{\text{Reconstruction Term}} \\
    &- \underbrace{ \mathbb{E}_{\mathcal{O} \sim \pi_\theta} \left[ D_{KL}(P_{true}(\mathcal{Z}|\mathcal{V}) ,||, P_{prior}(\mathcal{Z})) \right] }_{\text{Prior Penalty Term}},
    \tag{4}
\end{align*}
since both $P_{true}(\mathcal{Z}|\mathcal{V})$ and $P_{prior}(\mathcal{Z})$ are human prior distributions unrelated to $\pi_\theta$, the KL-divergence term is a non-negative constant. We focus on maximizing the reconstruction term:
\[
I(\mathcal{O}; \mathcal{Z} | \mathcal{V}) \ge \underbrace{ \mathbb{E}_{\mathcal{O} \sim \pi_{\theta}} }_{\rm CVIR} \underbrace{\left[ \mathbb{E}_{\mathcal{Z} \sim P^*(\mathcal{Z}|\mathcal{V})} [\log Q(\mathcal{Z} | \mathcal{O})] \right]}_{\rm DEDR}  \sim \mathcal{J}_{\rm DSPO}(\theta),
\tag{5}
\]
for the reconstruction term, $\mathbb{E}_{\mathcal{O} \sim \pi_{\theta}}$ denotes the ability for the MLLM to generate factually accurate descriptions, which refers to CVIR. Besides, $\mathbb{E}_{\mathcal{Z} \sim P^*(\mathcal{Z}|\mathcal{V})} [\log Q(\mathcal{Z} | \mathcal{O})]$ denotes the ability to use MLLM outputs to fit the true human distribution, which refers to DEDR. Thus, our proposed DSPO is fundamentally a variational lower bound of genuine human emotion.

\subsection{Experimental Setup}

\textbf{Datasets and Metrics.} We evaluate the emotion reasoning of MLLMs on three public benchmarks, \emph{i.e.,} EmoSet~\citep{yang2023emoset}, Emotion6~\citep{peng2015mixed}, and WebEmo~\citep{panda2018contemplating}. We evaluate DSPO on both in-domain and out-of-domain (OOD) benchmarks. Specifically, we leverage EmoSet/Emotion6 as the training source and the other two datasets as the external datasets. For each dataset, we use emotion accuracy as the evaluation metric. All reported accuracy metrics are computed via hard matching of the final $\backslash$boxed\{\} prediction against the ground-truth label.

\textbf{Base Model and Implementation Details.} We compare DSPO by two backbones: LLaVA-1.5-7B~\citep{liu2024improved} and Qwen2.5-VL-3B-Instruct~\citep{qwen2025qwen25technicalreport}, and with zero-shot inference, SFT, two reinforcement-learning baselines \emph{i.e.,} GRPO~\citep{shao2024deepseekmath} and DAPO~\citep{yu2026dapo}, and a SOTA method EMO-R3~\citep{fang2026emo} under two rollout budgets. We employ two evaluation settings: 1) EmoSet~\citep{yang2023emoset} for in-domain evaluation and Emotion6~\citep{peng2015mixed}/WebEmo~\citep{panda2018contemplating} for out-of-domain evaluation and 2) Emotion6 for in-domain evaluation and EmoSet/WebEmo for out-of-domain evaluation.
Following the standard hyper-parameter configurations established in prior GRPO-based works \cite{shao2024deepseekmath, fang2026emo}, we set the clipping parameter to $\epsilon = 0.2$ and the KL penalty coefficient to $\beta = 0.01$. For the ground-truth subjective distribution construction, we directly use the original multi-annotator emotion probability distributions for the Emotion6 dataset. For EmoSet and WebEmo, we leverage the Warriner \& NRC VAD lexicon ($\sim$54,800 lemmas) for static priors, and Mimo-v2 \cite{xiao2026mimo} for dynamic context captioning. The number of rollouts per group is set to $G = 4/8$, and the number of emotion candidates per rollout is $N = 5$ unless otherwise specified. The default reward coefficient is $\lambda_{acc}=1.0, \lambda_{fmt} = 0.1$, $\lambda_{cf} = 0.1$, and $\lambda_{div} = 0.3$. All experiments are conducted on $8\times$ A800 (80GB) GPUs.

\textbf{Distribution-based Evaluation.}
To intuitively quantify whether DSPO truly learns the authentic emotion distribution, we evaluate on four evaluation metrics based on emotion distribution. First, we use EmoSet as the training source and evaluate the models on the multi-annotator distribution labels of Emotion6. Furthermore, since DSPO and previous methods both generate only a single final emotion category per inference, we perform $M=50$ independent samplings for each image to approximate the distribution:
\[
{\rm Sample}_{M}(i)=\{y_i^1,y_i^2,…,y_i^M\},\quad q_i(c)=\frac{m_{i,c}+\alpha}{M+\alpha|C|},
\tag{6}
\]
where $y_i^t$ denotes the emotion category predicted for the image $i$ in the $t$-th sampling. $m_{i,c}$ is the number of times category $c$ appears in $M$ inferences, where $c\in C=\{Anger,Disgust,Fear,Joy, Sadness,Surprise\}$ is one of the six emotion labels in Emotion6. $\alpha$ is a smoothing coefficient introduced to prevent divergence collapse caused by emotional probabilities of zero. Besides, for the ground-truth emotion distribution, we suppose that for image $i$ there are $N_i$ multi-label annotations, and category $c$ receives $n_{i,c}$ annotations. The ground-truth emotion distribution could be expressed as $p_i(c)=\frac{n_{i,c}}{N_i}$.

\textbf{Distribution-based Metrics.} First, we consider using divergence-based metrics to measure the similarity between the predicted distribution and the ground-truth distribution. Specifically, we employ both KL and JS divergence due to the instability of KL divergence:
\[
    D_{KL}(p_i||q_i)=\sum_{c=1}^{|C|}p_i(c)log\frac{p_i(c)}{q_i(c)},
    \tag{7}
\]
\[
    D_{JS}(p_i.q_i) = \frac{D_{KL}(p_i||m_i)+D_{KL}(q_i||m_i)}{2},\quad m_i = \frac{p_i+q_i}{2},
\tag{8}
\]
besides, we also employ two entropy-based metrics to evaluate whether the uncertainty of the predicted distribution approximates the ground-truth:
\[
    H^{human}_i = -\frac{1}{log|C|}\sum_{c=1}^{|C|}p_i(c)logp_i(c),\quad H^{model}_i = -\frac{1}{log|C|}\sum_{c=1}^{|C|}q_i(c)logq_i(c),
\tag{9}
\]
\[
    \text{En-MAE}=|H^{model}_i-H^{human}_i|,
\tag{10}
\]
we will report the absolute entropy values of the models $H^{model}_i$ and their proximity to the ground-truth entropy $\text{En-MAE}$.

\subsection{Causal-grounded Emotional Thinking Template}
\begin{querybox}
    {\textbf{\textcolor{warmtitle}{Causal-grounded Emotional Thinking:}}} \\
    
    \verb|<step1>|\textit{Identify the core visual element (action, facial expression, object, or environment) that triggers the emotion. Finally, provide a bounding box coordinate in the format $[y_{min}, x_{min}, y_{max}, x_{max}]$ put in} \verb|\bboxed{}|.\verb|</step1>|
    
    \verb|<step2>|\textit{Reflect on the psychological state. Describe in detail how a human observer would emotionally resonate with this trigger, specifically expressing the valence and arousal of the feeling.}\verb|</step2>|
    
    \verb|<step3>|\textit{Synthesize the visual evidence and psychological reflection to generate $K$ distinct emotions. For each, provide an emotion label and a brief justification linking back to the visual trigger. Finally, provide an emotion list in the format $\{e_1,e_2,…,e_K\}$ put in} \verb|\list{}|.\verb|</step3>|

    \verb|<step4>|\textit{Based on the above analysis, choose the most appropriate option from the following emotional descriptions:}
    \begin{center}
        \textbf{[Emotion Set of Dataset]}
    \end{center}
    \textit{The chosen option MUST BE put in} \verb|\boxed{}|.\verb|</step4>|
\end{querybox}

\end{document}